# C3Net: interatomic potential neural network for prediction of physicochemical properties in heterogenous systems

Sehan Lee,*[a] Jaechang Lim[a] and Woo Youn Kim*[a,b,c]

Understanding the interactions of a solute with its environment is of fundamental importance in chemistry and biology. In this work, we propose a deep neural network architecture for atom type embeddings in its molecular context and interatomic potential that follows fundamental physical laws. The architecture is applied to predict physicochemical properties in heterogeneous systems including solvation in diverse solvents, 1-octanol-water partitioning, and PAMPA with a single set of network weights. We show that our architecture is generalized well to the physicochemical properties and outperforms state-of-the-art approaches based on quantum mechanics and neural networks in the task of solvation free energy prediction. The interatomic potentials at each atom in a solute obtained from the model allow quantitative analysis of the physicochemical properties at atomic resolution consistent with chemical and physical reasoning. The software is available at https://github.com/SehanLee/C3Net.

## Introduction

An accurate description of the free energy change associated with the transfer of a solute between different phases is crucial for understanding chemical and biological phenomena. The free energy change directly influences numerous physicochemical properties including solvation free energy, partition coefficients, and target binding affinity.[1,2] These properties are important for guiding the design of pharmacokinetic and pharmacologic studies in drug discovery. Therefore, much effort has been put into developing fundamental understanding and computational techniques for predicting the free energy change.[3-5]

The free energy change is mainly governed by interactions between the atoms comprising a system.[6] An interaction type of particular interest is a covalent bond which determines the characteristics of the atoms in a molecule. The other interactions governing the properties are noncovalent interactions which are responsible for the organization and stabilization of the system at an equilibrium state. A robust and accurate way to estimate these interactions is to use physics-based methods such as quantum mechanics and classical Newtonian mechanics. These methods have been widely used in calculating various properties of chemical compounds. However, they have limitations originating from the trade-off between prediction accuracy and computational efficiency.[7]

Many chemical and biological phenomena take place in large and complicated environments consisting of one or more phases with different physical and chemical properties. Therefore, both accuracy and efficiency for the description of the environments are critical to make a reliable and practical prediction model. The environment models can be roughly divided into two classes: explicit and implicit models.[4] The explicit models treat the environments in molecular or atomic resolution and compute every interaction between all pairs of components. Although the explicit models can provide a spatially resolved description of the molecules, they are computationally demanding as each molecule is simulated with appropriate force fields which may be developed or reparametrized on demand for different systems. The implicit approaches treat the environments as a continuum medium characterized by a single set of parameters on a macroscopic scale. Continuum approaches often describe well the thermodynamic aspects of the systems with relatively high efficiency compared to the explicit models.[8-10]

*In silico* approaches based on machine learning algorithms have been extensively applied to the chemical and biological problems. Recent studies demonstrate the possibility of the approaches to achieve both accuracy and efficiency for the prediction of the properties as an alternative to the physics-based methods. However, the high variability and complexity of the problems and limited experimental data available in public domain impede the development of reliable prediction models. Therefore, most studies have been focused to correlate the structure of a solute with its property without consideration of environmental effects, limiting their application to a single environment.[11-17]

Deep neural networks have become an attractive machine learning approach to complex problems with a large amount of data.[18,19] A key forte of the deep neural networks is to extract relevant features directly from raw data as input unlike conventional machine learning methods requiring handcrafted feature description or selection. Another characteristic of the deep neural network is to implicitly identify a nonlinear relationship between independent and dependent variables, which is important and challenging for the design and optimization of a prediction model for complex systems. However, the black box-like internal logic in the inference process prohibits scientific interpretation on prediction results.[20]

There are a few deep neural network models to predict the solvation free energies of druglike molecules in any pure solvent system.[21,22] These models use multiple techniques to extract structural features from the SMILES strings or molecular graphs of solute and solvent molecules as input information as well as to capture key solvent-solute interaction features from the structural features. Although these approaches make considerable progress in the prediction of the solvation free energy, there are some disadvantages that limit accuracy and generalizability of the models. First, the SMILES and molecular graphs do not encode 3D molecular structure information.

[a.] HITS incorporation, 124, Teheran-ro, Gangnam-gu, Seoul, Republic of Korea.
[b.] Department of Chemistry, KAIST, 291 Daehak-ro, Yuseong-gu, Daejeon 34141, Republic of Korea.
[c.] KI for Artificial Intelligence, KAIST, 291 Daehak-ro, Yuseong-gu, Daejeon 34141, Republic of Korea.
† Corresponding author Email: sehanlee@hits.ai / wooyoun@hits.ai

Conformational change of a solute affects solvent accessibility of solute atoms to interact with its environment and thus the free energy. Second, the description of the solvent environments with the structural features of a single solvent molecule as input limits the application of the approaches only to pure solvents. As considering that many important chemical and biological phenomena take place in complex systems comprising multiple phases and components, additional techniques are required for the proper representation of the properties of such complex systems.

In this work, we introduce a deep neural network model following fundamental physics laws for predicting physicochemical properties in complex systems including pure solvents, 1-octanol/water partitioning, and PAMPA, named C3Net (continuum-medium continuous convolutional neural network). C3Net describes systems in the 3D space as an atomistic solute embedded in continuum media whose average influences on the solute properties are described by a few parameters, which allows the application of C3Net to systems with multiple phases or components through the optimization of the parameters in training. All the properties are determined by the interactions of a solute with its environments and can be calculated from the sum of the interaction potentials over all solute atoms. The interaction terms are decomposed and computed by using distance-dependent continuous-convolutions over the atomic features of the solute and environment parameters. We employed multitask learning to train a model with a single set of weights for the target physicochemical properties, which increases the amount and heterogeneity of data for learning the systems and hence improve the model performance including its generalization ability in a data-deficient environment.

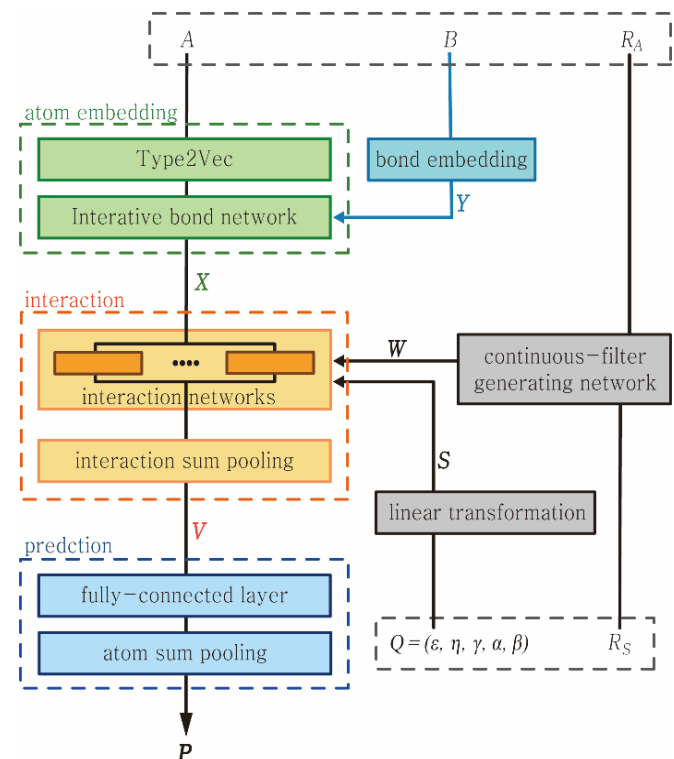

Fig. 1 The schematic architecture of C3Net.

## Methods

C3Net is designed for the prediction of molecular properties in complex systems by using a representation of interatomic potentials from a physics viewpoint. The model involves three sub-networks: the embedding block extracts atomic features of a solute, the interaction block calculates the decomposed interaction potential of the solute atoms with its environment, and the prediction block calculates the properties of the solute from the potentials. Fig. 1 shows an overview of the C3Net architecture.

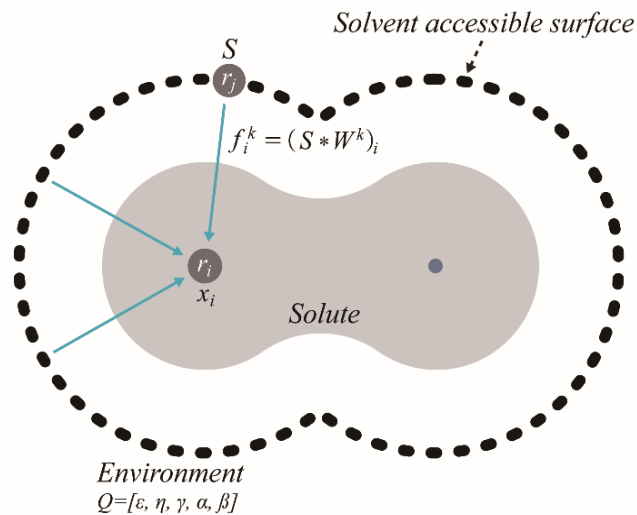

Fig. 2 The solute and environment are described as an assemblage of interacting compartments. Environment properties $Q$ are placed at the points on the solvent accessible surface.

**System Description** Three related systems of molecular transfer between or through different phases are considered. Solvation is an interaction of a solute with its surrounding solvent molecules leading to the stabilization of the solute in the solution system. Partition coefficients are related to the free energy of transfer of a solute between two immiscible phases. The PAMAP system consists of three phases and measures the ability of a solute to diffuse from a donor to an acceptor compartment separated by an artificial membrane. We focus primarily on the description of a solute with atomic resolution and represent the influence of the environment based on an approximate treatment as a homogeneous and isotropic continuum medium (Fig. 2). A solute molecule is treated as a chemical graph $G = (V, E)$ with vertices $V$ and edges $E$, denoting atoms $A$ and covalent bonds $B$, respectively. The features of a solute molecule is represented by atom and bond embeddings, $X = \{x_i | i \in V\}$ and $Y = \{y_{ij} | (i,j) \in E\}$, respectively, and positions $R_A = \{r_i | i \in V\}, R_A \in \mathbb{R}^3$. The environment system surrounding the solute is placed at $s$ points on the solute's solvent accessible surface (SAS), $R_S = \{r_i | i \in [1, \cdots, s]\}, R_S \in \mathbb{R}^3$. Since the focus of this work is primarily on the development of a deep neural network following physics laws, the representation of the environment is encoded by its physicochemical properties describing solvent effects on various solvent-solute interactions, including dielectric constant $\varepsilon$, refractive index $\eta$, surface tension $\gamma$, and hydrogen bond acidity $\alpha$ and basicity $\beta$, $Q = [\varepsilon, \eta, \gamma, \alpha, \beta]$.

**Data and Computational Setup** The training and validation of the model is performed on the combined set of experimental molecular properties in heterogeneous systems, including solvation free energies ($\Delta G_{solv}$), octanol-water partition coefficients (log $P_{ow}$), and apparent membrane permeability (log $P_{app}$). The 5,718 solvation free energies of 890 solutes in 103 solvents and the five solvent properties $Q$ are taken from Lee et al.[8] The 1-octanol-water partition coefficients of 14,043 structurally diverse compounds are obtained from the PhysProp database.[23] The membrane permeability of 665 compounds from the parallel artificial membrane permeation assay (PAMPA) are obtained from Fukunishi et al.[24] The environmental properties $Q$ of log $P_{ow}$ and log $P_{app}$ are initialized to the difference between the properties of water and 1-octanol and optimized in training. The dataset consists of neutral molecules containing only H, C, N, O, F, P, S, Cl, Br, and I atoms. We split the dataset randomly into eighty percent for training and twenty percent for external validation set. A maximum of 5 conformers is sampled for each molecule using the RDKit ETKDG method[25] combined with the universal force fields (UFF) energy minimization[26] to minimize the error originating from conformational changes in a solute upon interaction with different environments.[27] Each conformer is treated as an individual data point in training. The implementation of the neural networks is performed using PyTorch 1.8.1 deep learning framework.[28] The Adam optimizer with a learning late of 0.0001 and a batch size of 2 is used for training the model. Since the model is a regression problem, we use the mean squared error (MSE) as the loss function.

**Atom Embedding** The crucial part of the atom embedding is the atomic representation that extracts semantic relations between atoms with different characteristics. Word embedding is a vector representation of a word that captures the context of the word in a document and the semantic and syntactic relationship with other words from a large corpus of documents. Word2Vec is one of the most popular techniques to learn the word embedding using a shallow neural network.[29] Here we introduce Type2Vec which is an Word2Vec-inspired technique that treats atom types as "words" and compounds as "sentences" to convert an atom to a vector representing its physical and chemical properties in a context of neighboring atoms. In the first stage of atom embedding, atom and bond types are assigned to each atom and bond in a solute molecule, respectively. The atom and bond types are adapted from a net atomic charge calculation model, modified partial equalization of orbital electronegativity (MPEOE),[30] with a minor modification. The types are defined based on the atomic number, the type of hybridization, and the formal charge, and cover most of the known chemical space of drug-like molecules. Type2Vec uses 16-grams defined as a contiguous sequence of atom types within topological distance two. Type2Vec is trained on atom types and their contexts extracted from a large set of compounds obtained from the ChEMBL version 28[31] after removing the following compounds: (i) compounds containing more than one component or atoms other than H, C, N, O, F, P, S, Cl, Br, and I and (ii) compounds with molecular weights beyond a range of 200.0 to 600.0 g/mol. By applying the above criteria, about one million unique compounds are obtained, and their neutral and protonated states are generated using the Epik 2.7 implemented in Schrodinger.[32] While Type2Vec is a pre-trained network independent from C3Net, the bond type embeddings are initialized randomly and optimized during C3Net training. Properties of an atom in a molecule are influenced not only by its atomic features, but also by its chemical environment as a covalent framework allows charge rearrangement. The atom type embeddings encoded by Type2Vec are updated by an iterative bond network. At the $l$-th iteration, the bond network updates the atom embedding $X^l = \{x_i^l | i \in V\}$ using a residual connection.

$$x_i^{l+1} = x_i^l + q_i^l \quad (1)$$

The residual $q_i^l$ representing the influence of neighboring atoms on an atom $i$, $\{x_j^l | (i,j) \in E\}$, joined by covalent bonds, $\{y_{ij} | (i,j) \in E\}$, is calculated by pair-wise bond-filter convolutions.

$$q_i^l = (X^l * Y)_i = \sum_j x_j^l \circ y_{ij} \quad (2)$$

where "$\circ$" represents the element-wise multiplication. By using the iterative bond network, the influence of an atom propagates to the neighboring atoms through the covalent bonds.

**Solute-Environment Interactions** Given an environment representation $S$ derived from a linear transformation of the solvent properties $Q$, the potential field produced by the environment is decomposed into multiple terms by using parallel interaction networks with independent weights (Fig. 1). The $k$-th interaction network calculates a field at a solute atom $i$, $f_i^k$, using a continuous-filter convolution, which allows for rotation and translation invariance.

$$f_i^k = (S * W^k)_i = \sum_j^s S \circ W_{ij}^k \quad (3)$$

The convolution filter $W$ is the output of the continuous-filter generating network with gaussian radial basis functions.[33] The rotational and translation invariance is obtained by using the distance between the interacting solute atom $i$ and the environment point $j$, $d_{ij} = \|r_i - r_j\|$, as input for the filter network.

$$W_{ij}^k(r_i - r_j) = \exp\left[-\left(\frac{\|d_{ij} - \mu^k\|}{2\sigma^k}\right)^2\right] \quad (4)$$

The radial basis function parameters are expanded into the dimension of $X$ and optimized in training with initial values of 0 Å $\leq \mu \leq$ 8 Å and $\sigma$ = 0.1 Å. The interaction potential at the atom $i$, $V_i$, is then given by the sum of the decomposed components followed by the element-wise product.

$$V_i = x_i \circ \sum_k f_i^k \quad (5)$$

**Predictor** A property $P$ of a molecule is calculated in kcal/mol as the simple sum of atomic contributions, $p_i$, from a fully connected prediction network. The log $P_{ow}$ is directly proportional to the difference between the solvation free energies for a solute in water and 1-octanol solvents

$$\log P_{ow} = \frac{\Delta G_{water} - \Delta G_{1-octanol}}{RT \ln 10} \quad (6)$$

where $R$ is the universal gas constant and $T$ is the system temperature. The log $P_{ow}$ and log $P_{app}$ expressed in logarithmic units (base 10) are obtained by $P/RT\ln10$.

$$P = \omega \sum_i p_i, \quad \omega = \begin{cases} 1/RT\ln10, & \text{if } \log P_{ow} \text{ or } log\, P_{app} \\ 1, & \text{otherwise} \end{cases}$$

Since the embedding, bond, and interaction networks are independently applied to each atom and grid point by the convolution sum over neighboring atoms, the prediction of properties as the sum of individual contributions guarantees indexing invariance.

Table 1. Performance of C3Net for training and validation using single and multiple solute conformations

| | Multiple Conformations[a] | | | | | |
| --- | --- | --- | --- | --- | --- | --- |
| | Training | | | External Validation | | |
| System | $N_{data}$ | MAE | $R^2$ | $N_{data}$ | MAE | $R^2$ |
| $\Delta G_{solv}$ | 4587 | 0.103 | 0.999 | 1131 | 0.270 | 0.993 |
| log $P_{ow}$ | 11234 | 0.199 | 0.977 | 2809 | 0.290 | 0.951 |
| log $P_{app}$ | 532 | 0.186 | 0.947 | 133 | 0.361 | 0.851 |
| Total | 16353 | 0.171 | 0.997 | 4073 | 0.287 | 0.993 |
| | Single Conformation | | | | | |
| | Training | | | External Validation | | |
| System | $N_{data}$ | MAE | $R^2$ | $N_{data}$ | MAE | $R^2$ |
| $\Delta G_{solv}$ | 4582 | 0.282 | 0.993 | 1136 | 0.349 | 0.990 |
| log $P_{ow}$ | 11133 | 0.309 | 0.948 | 2798 | 0.342 | 0.930 |
| log $P_{app}$ | 533 | 0.492 | 0.720 | 129 | 0.504 | 0.717 |
| Total | 16248 | 0.307 | 0.993 | 4063 | 0.349 | 0.991 |

[a]mean of prediction values for conformers of a solute is used

## Results and Discussions

The C3Net architecture shares all the weights and parameters across its network to improve generalization performance. The overall performance of C3Net is summarized in Table 1 and Fig. 3. The results indicate that the model reproduces the physicochemical properties very well for diverse compounds in complex and heterogeneous systems. The $R^2$ of the model with multiple solute conformations is 0.997 over the entire training set. It is noted that one of the focuses of this study is on the minimization of the error originated from structural changes upon dissolution in different media. Our results of the parameterizations for the same data set with a single conformation indicate that the use of multiple conformations can yield better performance than that with single conformation, where MAEs for the validation set are improved from 0.349 to 0.270 kcal/mol, 0.342 to 0.290 log units, and 0.504 to 0.361 log units in solvation, 1-octanol-water partitioning, and PAMPA systems, respectively. Hereafter, we use results with the multiple conformations. Table 1. Performance of C3Net for training and validation using single and multiple solute conformations

The performance of C3Net on each system applied to the external validation set is shown in Fig. 4. The MAEs for the systems are within 0.5 kcal/mol and only one solvation system in perfluorobenzene has MAE of 0.786 kcal/mol. The experimental and predicted values are highly correlated in all systems over the whole MAE ranges; the lowest $R^2$ values are 0.549 in nitrobenzene, 0.841 in PAMPA, and 0.883 in n-methyl-2-pyroolidone, and $R^2$s for the other systems are higher than 0.900. The results indicate that the model is fairly generalized to heterogeneous and complex systems.

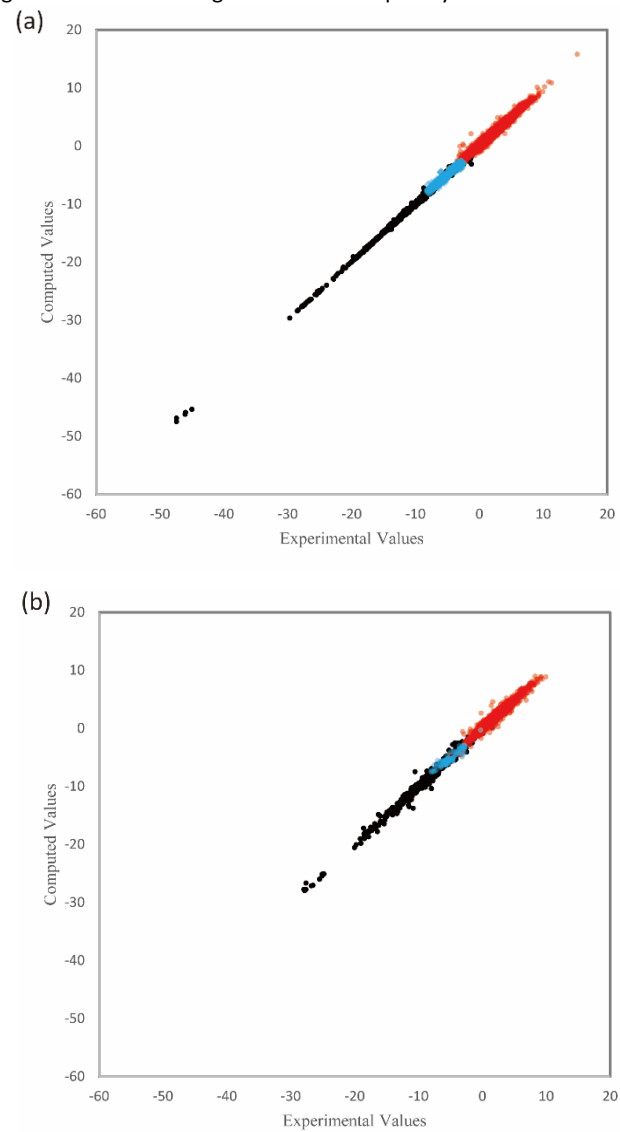

Fig. 3 Scatter plots of experimental and predicted values for the training (a) and validation data (b). The black, red, and blue dots represent the solvation free energies, octanol-water partition coefficients, and PAMPA permeability, respectively.

The solvation free energies of small molecules in physically diverse solvents are reproduced well. The MAEs for the training and validation sets are 0.103 and 0.270 kcal/mol, respectively. We further compared our results on the solvation free energies with other state-of-the-art methods based on quantum chemistry, including conductor-like screening model (CSOMO) and solvation model (SM), and recurrent neural network, deep learning model for solvation free energies in generic organic solvents (Delfos). Among the different versions of the methods, the versions with the best performance were compared, COSMO-RS with BP8 functional and TZVP basis, SM12CM5 with B3LYP functional and MG3S basis, and Delfos with BiLSTM architecture. Although overall those methods are highly accurate within 1.0 kcal/mol for both non-aqueous and aqueous solutions, our physics-informed architecture shows much better prediction than those in terms of MAE (Table2). C3Net achieves 0.10 and 0.27 kcal/mol for the training and validation sets of organic solvents, respectively, which is comparable to 0.24 kcal/mol of Delfos and superior to 0.41 and 0.54 kcal/mol of COSMO and SM, respectively. C3Net provides much better prediction for water solvent, where the MAEs of C3Net are 0.09 and 0.333 kcal/mol for the training and validation sets, respectively, and those of the other models are in a range of 0.52-0.77 kcal/mol. This is the most accurate solvation free energy calculation in the literature to the best of our knowledge.

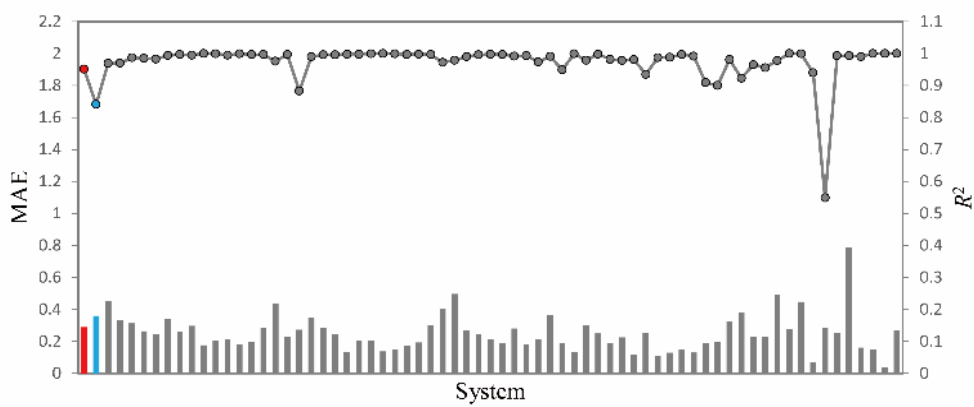

Fig. 4 Performance, MAE and $R^2$, for an external validation data. MAE (left y-axis, bars) and $R^2$ (right y-axis, circles) are plotted for each system. The grey, red, and blue represent the solvation free energies, octanol-water partition coefficients, and PAMPA permeability, respectively. The systems are sorted by the number of validation data in a descending order.

Table 2. Comparisons between solvation models for diverse solutions. The error metric is MAE and kcal/mol. The values in the parentheses denote the external validation results.

| Solvent | Method | $N_{data}$ | $N_{solvent}$ | MAE (kcal/mol) |
|---|---|---|---|---|
| Aqueous | C3Net | 281 (71) | 1 | 0.09 (0.33) |
|  | COSMO-RS/BP86/TZVP[9] | 274 | 1 | 0.52 |
|  | SM12CM5/B3LYP/MG3S[10] | 374 | 1 | 0.77 |
|  | Delfos/BiLSTM[22] | 374 | 1 | 0.64 |
| Non-aqueous | C3Net | 4306 (1060) | 102 | 0.10 (0.27) |
|  | COSMO-RS/BP86/TZVP[9] | 2072 | 90 | 0.41 |
|  | SM12CM5/B3LYP/MG3S[10] | 2129 | 90 | 0.54 |
|  | Delfos/BiLSTM[22] | 2121 | 90 | 0.24 |

A large set of structurally diverse compounds with a wide range of log $P_{ow}$ (-5.08 to 15.4) is predicted with high accuracy. The MAEs for the training and validation sets are 0.199 and 0.290 log units, respectively, which are within experimental uncertainties of 0.385 log units.[23] C3Net is applied to predict the log $P_{ow}$ values of 87 peptides[34] which are not used in the model development. The dataset includes 76 blocked, six unblocked, and five cyclic neutral peptides and consists of one to eight natural and synthetic amino acids. In Figure 5, the calculated log $P_{ow}$ values of the peptides are plotted against experimental values. The MAE and $R^2$ of the neutral peptides were 0.438 log units and 0.878, respectively. The model predicts the values well for most of the 87 peptides; however, eight ionizable peptides, including melanotan-II, sandostatin, and six unblocked peptides, are systematically over-predicted by 1.25 to 2.66 log units, as observed in other studies.[35] If the ionizable peptides are removed, the MAE and $R^2$ are 0.316 log units and 0.953, respectively.

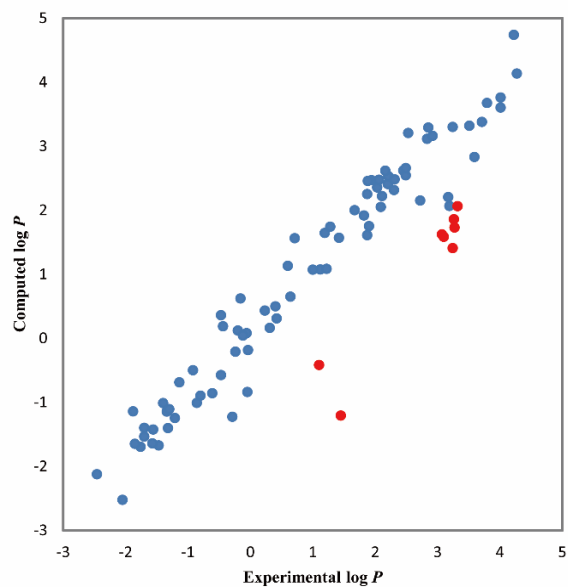

Fig. 5 The experimental octanol-water partition coefficients vs. calculated values for 87 blocked, unblocked, and cyclic peptides. The red dots represent the ionizable peptides.

In this work, the PAMPA system is treated as three layers of distinct continuums, and we assume that the passive membrane permeability depends on the solvation free energy difference between the continuums. Again, C3Net uses a set of weights and parameters to learn interactions in heterogeneous systems and only five parameters $Q$ describing characteristics of the PAMPA and 1-octanol-water partitioning systems are optimized. C3Net provides highly accurate results with MAEs of 0.186 and 0.361 log units for the training and validation sets, respectively, which are within experimental uncertainties, 0.401 log units.[24]

Within the C3Net architecture, interaction potentials are calculated at each atom in a solute molecule and a physicochemical property is represented as the sum of the potentials. To improve the understanding of physics behind the properties in complex systems, quantitative analysis of atomic contributions to solvation free energy is performed. Figure 6 shows the atom-wise breakdown of the solvation free energies of n-pentane and pentanoic acid in water, 1-octanol, and n-hexadecane in units of kcal/mol. The aliphatic carbon and hydrogen atoms of n-pentane in the polar water solvent decrease the solvation free energy due to their hydrophobic nature, and the more solvent exposed terminal $CH_3$ groups have the more positive values. The aliphatic carbon atoms have positive contributions in the hydrophobic solvent, n-hexadecane, and have values between water and n-hexadecane in 1-octanol which has both a polar hydroxyl group and a hydrophobic carbon chain. The contribution of polar carboxylic acid atoms in pentanoic acid increases as the solvent polarity increases, but hydrophobic carbon atoms show the opposite trend as shown in n-pentane. Note that the carboxylic acid group increases the solvation free energies even in the hydrophobic n-hexadecane solvent and the solvation free energy of polar hexanoic acid is higher than that of hydrophobic n-hexane in the hydrophobic n-hexadecane solvent, which are -4.61 and -2.95 kcal/mol, respectively.

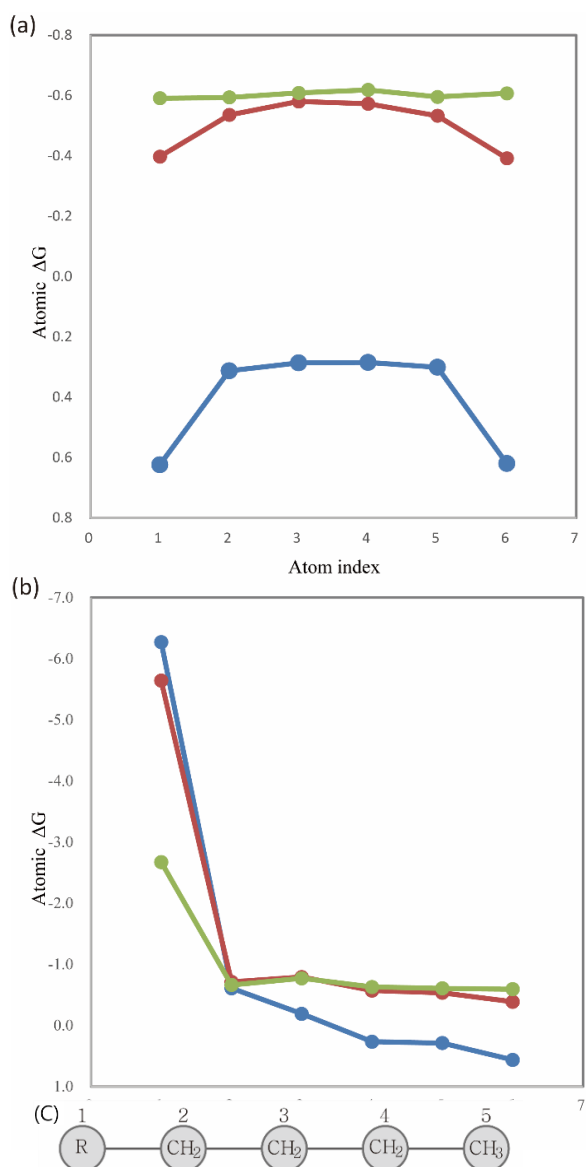

Figure 6. Atom-wise contributions to the solvation free energies of n-hexane (a) and hexanoic acid (b) in water (blue), 1-octanol (red), and n-hexadecane (green). Atom index is shown in (c), in which R is $CH_3$ and carboxylic acid in n-hexane and hexanoic acid, respectively.

## Conclusions

We have described a deep neural network, C3Net, for the prediction of physicochemical properties of a solute in heterogeneous systems including solvation in diverse solvents, 1-octanol-water partitioning, and PAMPA. C3Net aims to provide a practical framework by means of a proper description of the systems and effects of solute and environment on the properties. In the model, various solute-environment interactions are computed using parallel interaction networks, which follows fundamental physics including description of heterogeneous systems in addition to invariance to atom indexing, rotation, and translation. The effects of the solute on the interactions are represented by a fixed size vector encoded by Type2Vec and a bond network, and the complementary effects of the environment on these interactions are reflected with a set of macroscopic properties. This approach gives us more intuitive and interpretable insights into quantitative contributions of interaction potentials at solute atoms to a physicochemical property, which is valuable in the rational design of molecules with desired properties.

One of the largest advantages of our approach lies in its high generalization ability to larger and more complex systems. Given the pre-trained C3Net network, the only parameters needed for the prediction of a property in a new system are a set of properties describing the environment, $Q$, which can be optimized with a small size of experimental data. Possible applications we can consider are the prediction of solvation free energies in solvent mixtures and ionic liquid.

## Data availability

All data sets used in this study are publicly available. A detailed protocol for the reproducible collection and preprocessing of the data utilized in this work is provided in the Methods section. The code used for model training can be accessed at https://github.com/SehanLee/C3Net.

## Author Contributions

Sehan Lee: conceptualization, methodology, investigation, data curation, formal analysis, writing - original draft; Jaechang Lim: formal analysis, resources, writing – review & editing; Woo Youn Kim: supervision; project administration, writing – review & editing.

## Conflicts of interest

There are no conflicts of interest to declare.

## Acknowledgements

This work was supported by the Tech Incubator Program for Startup (TIPS) funded by the Ministry of SMEs and Startups (MSS, Korea) (S3031674).